\documentclass[conference]{IEEEtran}
\IEEEoverridecommandlockouts
\usepackage{cite}
\usepackage{amsmath,amssymb,amsfonts}
\usepackage{algorithmic}
\usepackage{url}
\usepackage{todonotes}
\usepackage{graphicx}
\usepackage{textcomp}
\usepackage{caption}
\usepackage{subcaption}
\usepackage{xcolor}
\usepackage{tablefootnote}
\usepackage{threeparttable}
\usepackage{multirow}
\usepackage[T1]{fontenc}
\usepackage{booktabs}
\newcommand{\tabitem}{~~\llap{\textbullet}~~}

\usepackage{hyperref}
\def\BibTeX{{\rm B\kern-.05em{\sc i\kern-.025em b}\kern-.08em
    T\kern-.1667em\lower.7ex\hbox{E}\kern-.125emX}}
 
\newenvironment{myitemize}{
\begin{itemize}
 \setlength{\itemsep}{1pt}
 \setlength{\parskip}{0pt}
 \setlength{\parsep}{0pt}
}{\end{itemize}}

\newenvironment{myenum}{
\begin{enumerate}
 \setlength{\itemsep}{1pt}
 \setlength{\parskip}{0pt}
 \setlength{\parsep}{0pt}
}{\end{enumerate}}

\begin{document}

\title{ Scalable Fact-checking with Human-in-the-Loop\\
}

\author{Jing Yang$^1$, Didier Vega-Oliveros$^1$, Tais Seibt$^2$, and Anderson Rocha$^1$\\\\
$^1$ Artificial Intelligence Lab. (\url{Recod.ai}), Institute of Computing, University of Campinas, SP, Brazil \\
$^2$ Department of Journalism, 
Universidade do Vale do Rio dos Sinos (Unisinos), 
Porto Alegre, RS, Brazil}

\maketitle


\begin{figure}[b]
\vspace{-0.3cm}
\parbox{\hsize}{\em
WIFS`2021, December 7 – 10, 2021, Montpellier, France.
XXX-X-XXXX-XXXX-X/XX/\$XX.00 \ \copyright 2021 IEEE. 
}\end{figure}

\begin{abstract}

Researchers have been investigating automated solutions for fact-checking in a variety of fronts. However, current approaches often overlook the fact that the amount of information released every day is escalating, and a large amount of them overlap. Intending to accelerate fact-checking, we bridge this gap by grouping similar messages and summarizing them into aggregated claims. 
Specifically, we first clean a set of social media posts (e.g., tweets) and build a graph of all posts based on their semantics; Then, we perform two clustering methods to group the messages for further claim summarization. We evaluate the summaries both quantitatively with ROUGE scores and qualitatively with human evaluation. We also generate a graph of summaries to verify that there is no significant overlap among them.
The results reduced 28,818 original messages to 700 summary claims, showing the potential to speed up the fact-checking process by organizing and selecting representative claims from massive disorganized and redundant messages. 

\end{abstract}

\section{Introduction}
As misinformation becomes a growing concern to the public, news fact-checking organizations are also proliferating. However, the generation and spreading speed of the former is much faster than the latter. To fight misinformation, many automated solutions for fact-checking have been proposed. A typical pipeline for automated fact-checking usually consists of four steps: claim check-worthiness detection, evidence retrieval, evidence selection, and veracity verification~\cite{hassan2017toward,thorne2018fever}. Some works bring about additional flavors to the  pipeline by: 1) matching a claim with verified claims before checking to avoid repetitive work~\cite{shaar2020known}; and 2) generating explanations after the veracity verification to make the result more reliable~\cite{atanasova2020generating}. 

Automated fact verification has received most attention in the literature. However, human fact-checkers often do not trust results from automated solutions~\cite{nakov2021automated}. The reason is that automated methods are error-prone, and incorrect fact-checking could seriously damage fact-checking organizations' reputations. Instead, what fact-checkers seek from automated methods is to scale-up manual fact-checking's speed. Indeed this is essential in fact-checking, as every day, billions of messages are posted on social media~\footnote{\url{https://tinyurl.com/nt4pxa6a}} and misinformation is ever increasing. Therefore, fact-checkers cannot debunk every single post. Till now, most researchers have tried to handle this issue by checking if a post is worth-checking~\cite{clef2018checkthat:task1} to reduce the number of claims. However, this may not be enough as social media messages are noisy, and check-worthiness detection needs manual labeling which is bias prone. 

Having this in mind, we approach this problem under an unsupervised perspective. We notice that posts from social media overlap extensively; most of them are slight modifications or paraphrases of other posts. To exploit this observation, this paper proposes to assist human fact-checkers by grouping semantically similar claims together and summarize them into single key claims. The grouping stage consists of separating posts into distinct claims. The summarization stage aims at reducing redundancy and formulating an informative and representative claim. This is the first work that addresses grouping and summarizing semantical messages to scale up fact-checking to the best of our knowledge. Therefore the contributions of this work are summarized as follows:

\begin{myenum}
    \item We adopt two clustering methods, and both show that we can group semantically similar posts related to one claim or similar claims together.
    
    \item We adopt four methods for posts summarization (extractive and abstractive ones); The summaries are representative and reduce the redundancy in raw short messages.
    
    \item We generate a graph of summaries to verify the clustering and summarization methods; the graph shows that the summaries are well-separated.
    
    \item The validation brings humans back to the loop to assess claims worthiness to a fact-checker specialist. 
    
\end{myenum}

\section{Related Work} \label{review}

\subsection{Check-worthiness Detection}

Check-worthiness detection is related to our work and serves a similar purpose --- to reduce the number of claims to be checked. Currently,  most check-worthiness detection work focus on checking claims related to political debates. The well-known ClaimBuster~\cite{hassan2017toward} extracts, ranks, and identifies essential factual claims from presidential debates sentences. CheckThat! Lab.~\cite{clef2018checkthat:task1} has hosted since 2018 an open detection task of check-worthy claims. The goal is to give check-worthiness scores to a list of sentences. The Prise de Fer's team~\cite{zuo2018hybrid} proposed a hybrid model with various sentence representations, including both syntactic and semantic features. The Copenhagen team~\cite{hansen2019neural} extracted sentence features by a Recurrent Neural Network (RNN) with Gated-Recurrent Units (GRU) memory units. They used contrastive sampling to select sentence pairs further trained for check-worthiness prediction. Our goal differs from check-worthiness detection; rather than predicting if a claim is check-worthy we provide human experts meaningful summarized claims to facilitate the arduous fact-checking task.

\subsection{Social media message summarization}

Social media post summarization brings another branch of related works. These studies are usually related to event/disaster discovery. For example, Rudra et al.~\cite{rudra2016summarizing} reported a framework to summarize messages from Twitter. Their method comprises two stages. The first selects essential tweets from the whole set. The second combines selected tweets and generates a new message by maximizing tweets' informativeness and avoiding redundancy. Another example is the systematic review of summarization on tweets/microblogs for emergency events reported by Dutta et al.~\cite{dutta2019summarizing}. The authors analyzed eight summarization methods and showed that different methods generated very different summaries. 

Thus, although some tweets summarization methods have been proposed, there is still much room for improvement. To encourage research on this field, recently, Dusart et al.~\cite{dusart2021issumset} proposed a large dataset for tweet summarization of events named ISSumSet. This dataset contains 122 events, including COVID-19-related ones. Each event has various related tweets, and the tweets are labeled with different types and levels of importance. Tweets summarization is a challenging and vital task. Many messages are posted online daily, not to mention that they are noisy and multilingual. In fact-checking, we face similar challenges, and the summarization of posts to generate relevant claims may turn the task much more scalable.
\section{Task Formulation}
\begin{figure*}[h]
        \centering
        \includegraphics[width=0.8\textwidth]{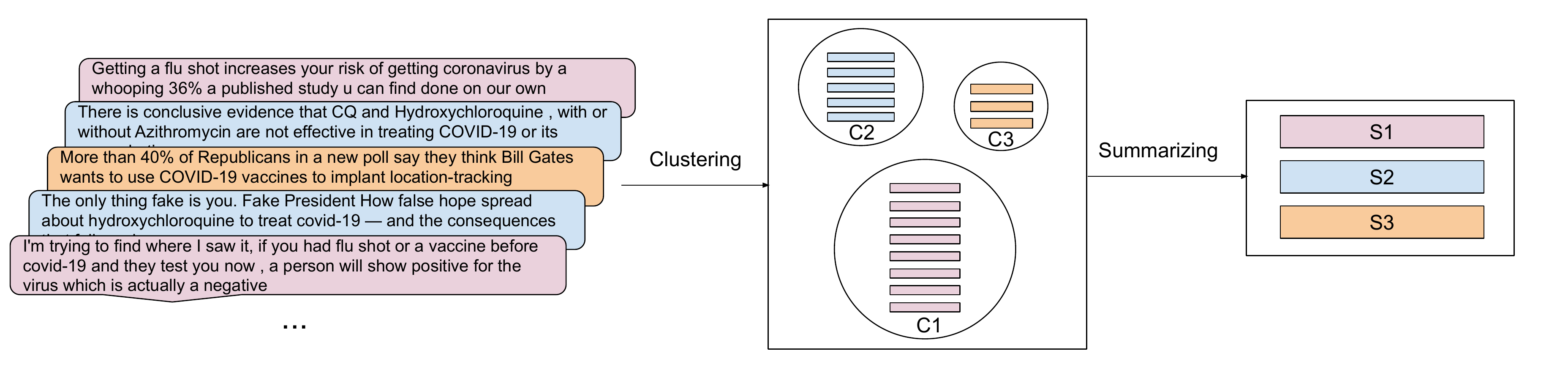}
        \caption{Unsupervised social media posts summarization pipeline. Given a set of posts, we perform two steps: semantic clustering and summarizing/claim generation. The first step groups posts into different clusters and ranks them based on the number of meaningful posts in each cluster. The second step summarizes messages from each cluster to generate an informative summary.}
        \label{f:pipeline}
    \end{figure*}

We define our task as follows: 
\textit{given a set of social media posts, separate them into different groups based on their semantics; then, summarize all posts for each group to generate an informative claim that can represent the group to aid fact checkers later on.} Figure~\ref{f:pipeline} shows the general pipeline for the task, consisting of two main steps: semantic clustering and content summarization/claim generation. 

\subsection{Short Message Aggregation}

Aggregation seeks to group short messages related to a single claim together. This group should include duplicated, near-duplicated, and paraphrased posts with same or opposite sentiments towards a claim. This is challenging because clustering is unsupervised; it is difficult to define the boundary for a short message to belong to one group or another.

State-of-the-art transformer-based language models are good at capturing semantic meanings of words. In our case, we need to capture semantic meaning of sentences, therefore we leverage Sentence-Transformers\footnote{\url{https://tinyurl.com/w4de7tvv}}\cite{reimers2019sentence} for short messages embeddings as input for clustering. 

There are different ways to perform aggregation or clustering. A standard clustering method is k-Means. However, it is not suitable in our case as each short message embedding has a dimension of at least 512, which makes k-Means relatively slow and, most importantly, it requires us to pre-define a specific number of clusters. Therefore, we adopt and compare two methods: Agglomerative clustering\footnote{\url{https://tinyurl.com/vr3p25w3}} and Leiden community detection\cite{traag2019louvain} as they do not require the establishment of the number of clusters beforehand.

    \subsubsection{\textbf{Agglomerative clustering}} 
    
    Hierarchical clustering groups feature points based on their dissimilarity. The method starts with each point as a cluster and merging two clusters into one if their dissimilarity value is below a decision cutoff. This method is helpful  because the number of clusters is unknown; we can control the decision cutoff to have smaller or larger clusters. First, we calculate a similarity matrix $S$ of short message embeddings for the initial dissimilarity values, then provide $1-S$ as the dissimilarity matrix. For the linkage criteria determining how the dissimilarity is calculated between two clusters, we choose the average dissimilarity between any two points in the two clusters.
    
    \subsubsection{\textbf{Leiden community detection}}
    
    Leiden community detection is a graph-based clustering method that finds the best community partition in a graph~\cite{traag2019louvain}. It improves the convergence time of the Louvain algorithm with a smaller computational footprint, providing partitions focused on the micro-patterns of the communities that maximize the graph modularity. Formally, the graph $G(N,L)$ is formed by the set of nodes $N$ representing each short message; and the set of links $L$, which represent the similarity weight between nodes. The construction process from the short message embedding to the graph calculates the similarity matrix among the vectors and then applies the $\epsilon$-neighborhood method~\cite{Luxburg2014}. 

For both methods, we use the cosine similarity to compute the similarity matrix of all post representations. As both methods require a decision cutoff for similarities, we denote the threshold $\delta$ and $\epsilon = \delta$ as the $\epsilon$-graph construction parameter.  

\subsection{Short message Summarization}

For each cluster, our goal is to summarize its short messages to generate a claim. We leverage two types of summarization. 
    \subsubsection{\textbf{Extractive summarization}}
    
    For extractive summarization, we aim to select a representative short message from each cluster. In particular, we construct an $\epsilon$-graph for each cluster and use centrality measures to rank the short messages in the cluster and select the most central one as the summary. The idea of using centrality measures is that central nodes are usually the more influential or representative in the graph~\cite{vega2019multi}. We adopted two methods: the Degree Centrality (DG) and the Multi-Centrality Index (MCI)~\cite{vega2019multi}. The DG counts the total number of input/output connections of the nodes, and nodes with the highest DG centrality are known as hubs. The MCI considers multiple measures for finding the most relevant message in the cluster. In this work, we consider the Degree, PageRank, and Betweenness centrality, along with the number of reposts and likes of each message, for calculating the MCI.  
    
    \subsubsection{\textbf{Abstractive summarization}}
    
     For the abstractive summarization, we adopt two state-of-the-art transformer-based language models: BART\cite{lewis2020bart} and T5\cite{raffel2020exploring} to generate a summary. The challenge of these two models is that the models' maximum input length is not long enough to fit all short messages in some clusters. Before feeding all the messages to the summary process, we remove duplicates and near-duplicates from each cluster to deal with this problem. We perform the agglomerative clustering with a higher similarity threshold within each cluster. Afterward, we have more sub-clusters in each cluster, and each sub-cluster contains only messages that are duplicates and near-duplicates. We randomly select one message from each sub-cluster as messages in the same sub-cluster can be treated as equivalent.
    
\section{Evaluation and Analysis}
\subsection{Dataset}

For the evaluation of our proposed pipeline, we adopt MM-COVID\cite{li2020mm}, a fake news detection dataset. Each news article in this dataset is accompanied by social media context: tweets, retweets, and replies. Here, we only use tweets content, as retweets are duplicates of tweets, and replies can be less related to the claim itself. We adopted this dataset because we can use its labels for evaluation, as each news piece has a claim summarizing the news content. This news summary can be treated as the ground-truth for our short messages summary. We emphasize that although the dataset was proposed for supervised learning on text classification, we only use the labels for evaluation, i.e., our methods perform unsupervised learning all the time. Our code will be available at: \url{https://github.com/jingyng/scalable-fact-checking}.

Through the Twitter API\footnote{\url{https://developer.twitter.com/en/docs/twitter-api}}, we collected 92,070 tweets associated with 2,227 news articles (around 12\% tweets were removed from Twitter at the time of collection). Out of all tweets,  48,074 tweets (52.2\%) associated with 1,092 news articles are in English. In this work, we consider only English tweets, but our pipeline can be easily adapted to other languages as long as trained language models are available. After collecting all the tweets, we pre-processed them by removing duplicated (tweets with the same id), user mentions, URLs, hashtags, and emojis.

One challenge of using this dataset for evaluation is that there are some mismatches between news claims and tweets content, i.e., a tweet associated with one news piece is not related to its news content. We show one example here:

\textbf{News claim}: \textit{Coronavirus is caused by 5G.}, 

\textbf{Tweet content}: \textit{Recently, we have also had some misinterpret some CDC data related to deaths from COVID-19. Without a doubt, we know coronavirus has caused more than 400 deaths in Utah and over 177,000 in the United States.}

The example shows a tweet content not related to the news claim. This hinders us from using news claims as gold summaries for a tweet cluster. Therefore, we need to process the dataset further. We remove tweets less relevant to a news claim based on a relevance decision cutoff $\theta$. This step is only necessary to evaluate summarization and does not need to be performed in real cases.

For calculating the relevance between tweets and their news summary, we rely upon BERTscore\cite{zhang2019bertscore}, as it has shown better performance than cosine similarity in~\cite{hossain2020covidlies}. We adopt the model (roberta-large) for BERTscore and normalize the score\footnote{\url{https://github.com/Tiiiger/bert_score}}. After calculating the relevance between tweets and news summaries, we remove all tweets irrelevant to its news summary, with the threshold $\theta = 0.1$. We also filter out messages with less than 4 words, given that they do not contain meaningful information. After this process, we have 28,818 remaining tweets associated with 959 news original articles. 

\subsection{Clustering Evaluation}
As we do not have ground-truth cluster labels, we rely upon the Silhouette coefficient metric\footnote{\url{https://tinyurl.com/3xfvsbck}} to evaluate the clustering methods. This metric ranges from -1 to 1, with a higher value indicating a better-defined cluster with less overlap among clusters. To compare the clustering results, we consider two factors: embedding models and clustering methods. 

    \subsubsection{\textbf{Comparison of embedding models}}
    
    A good embedding model is essential in clustering; it should map semantically similar messages closer in their feature representation space. To compare different embedding models, we set the decision cuttoff $\delta = 0.85$ and the clustering method to be Leiden community detection. Table~\ref{tab:emb} shows the results. 
    
    \begin{table}
    \centering
    \caption{Clustering results comparison between different embedding methods.}
    \begin{tabular}{lll}
    \hline
    \begin{tabular}[c]{@{}l@{}}\textbf{Embedding  model}\end{tabular} & \begin{tabular}[c]{@{}l@{}}\textbf{\# of clusters}\end{tabular} & \begin{tabular}[c]{@{}l@{}}\textbf{Silh.  Coef.}\end{tabular} \\ \hline
    paraphrase-distilroberta-base-v2                           & 701                                                           & 0.76                                                            \\
    paraphrase-mpnet-base-v2                                   & 677                                                           & 0.73                                                            \\
    paraphrase-MiniLM-L6-v2                                    & 707                                                           & 0.75                                                            \\
    nli-mpnet-base-v2                                          & 739                                                           & 0.79                                                            \\
    nli-roberta-base-v2                                        & 705                                                           & 0.79                                                            \\
    digitalepidemiologylab/covid-twitter-bert-v2               & 732                                                           & 0.76                                                            \\ 
    cardiffnlp/twitter-roberta-base               & 1                                                                      & -- \\ \hline
    \end{tabular}
    \label{tab:emb}
    \end{table}

The clustering performance varies but all embedding models (except for~\textit{cardiffnlp/twitter-roberta-base}) lead to reasonable performances. Surprisingly, embedding model ~\textit{cardiffnlp/twitter-roberta-base} only resulted in one big cluster, although it was pre-trained with tweets. All other models comprise about 700 clusters, less than the number of news, 959 indicating that some news claims are similar to each other.
    
    \subsubsection{\textbf{Comparison of clustering methods}}
    
\begin{figure}[t]
    \centering
        \includegraphics[width=0.37\textwidth]{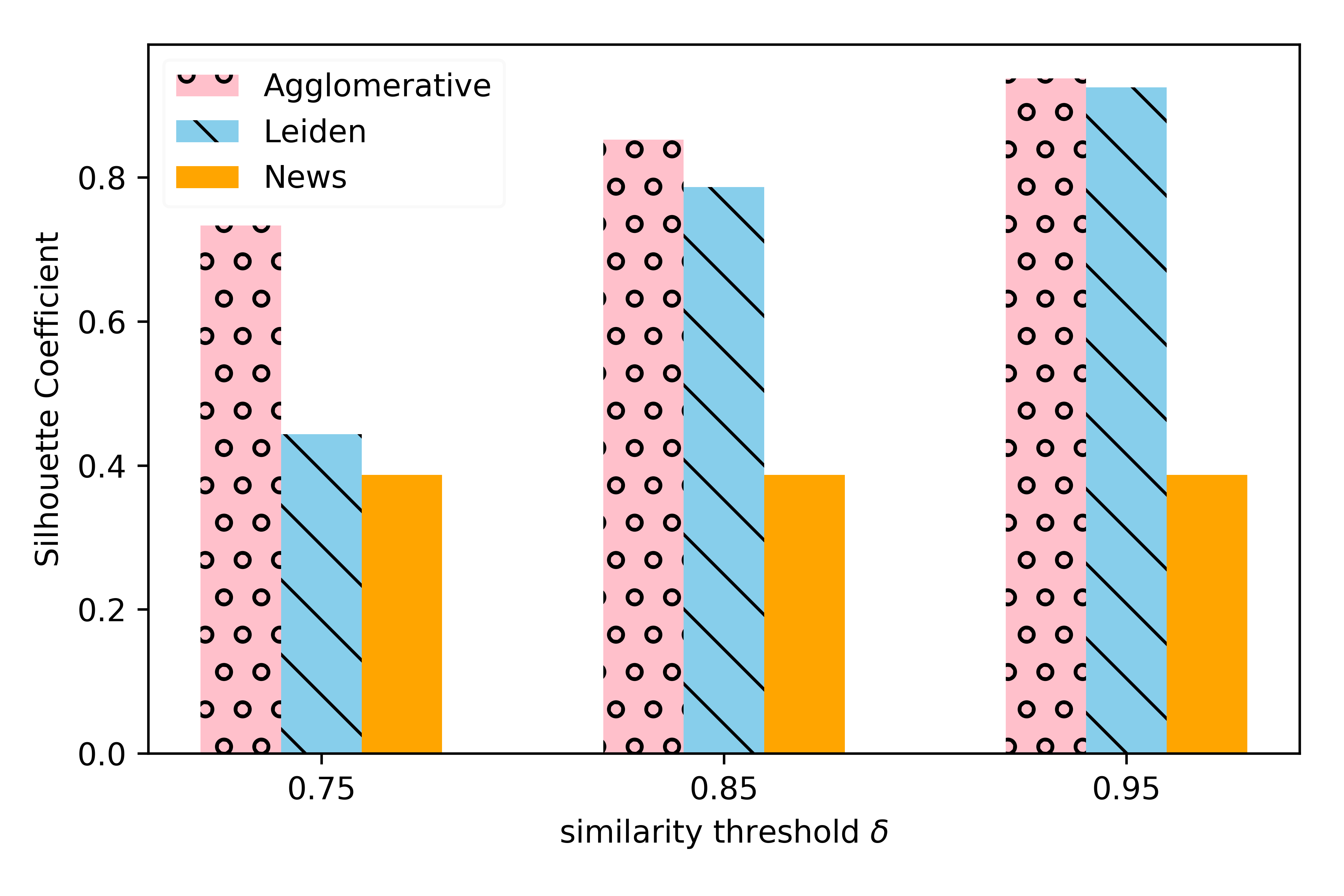}
    \caption{Clustering results varying the similarity threshold $\delta$.}
    \label{fig:dist}
    \end{figure}

    As previously mentioned, we adopt two clustering methods: Agglomerative clustering and Leiden community detection. As a comparison, we also consider the original posts clustering separated by news (i.e., each news corresponds to one cluster of posts). 
    In Figure \ref{fig:dist}, we vary similarity threshold $\delta$ to compare clustering performance. We fixed the embedding model~\textit{nli-roberta-base-v2} as it performed best (see Table \ref{tab:emb}). The Silhouette coefficient increases when $\delta$ increases. When $\delta$ is close to 1, Agglomerative clustering and Leiden clustering methods yield about the same results because when $\delta$ is high they are only grouping near-duplicated tweets together. This indicates that the Silhouette coefficient can only partially evaluate clustering results, as we want to cluster posts that are semantically similar to each other, not just posts that contain similar words.

\subsubsection{\textbf{Distribution of news in clusters}}

To check if the method is indeed clustering posts related to one news claim, we analyze news in each cluster. The percentage of clusters with only one associated news claim for agglomerative and Leiden are 95.15\% and 93.34\%, respectively. We randomly examine one cluster with more than one associated news piece (agglomerative method) to see if the news claims are similar. One example of news claims in a cluster is the following:

\begin{myitemize}
    \item \textit{U.S. President Donald Trump or presidential candidate Joe Biden referred to the novel coronavirus pandemic as a time when ``people are dying that have never died before.''}
    \item \textit{ Donald Trump said about coronavirus, ``People are dying who have never died before.''}
    \item \textit{Referring to the ongoing COVID-19 pandemic, U.S. President Donald Trump said, ``People are dying today that have never died before.''}
\end{myitemize}

The three news claims are indeed related to one claim. This also explains why the number of clusters (agglomerative: 804, Leiden: 705) is less than the number of news articles (959).

\subsection{Summarization Evaluation}
After aggregating the posts, we perform two types of summarization: extractive and abstractive summarization. For the quantitative evaluation of the summarization results, we use the F1-score of ROUGE-1, ROUGE-2, and ROUGE-L metrics\footnote{\url{https://tinyurl.com/3e5t8tpz}}.  ROUGE scores are common metrics for text summarization tasks. Given a generated summary and a reference summary, ROUGE-1 and ROUGE-2 measure the overlap of unigram and bigram, respectively, and ROUGE-L measures the Longest Common Subsequence (LCS) between them.
We also use BERTscore (pre-trained with roberta-large model) to measure the semantic similarity between the generated summary and the ground-truth summary. We use the average summary length to indicate the informativeness of summaries.

    \subsubsection{\textbf{Comparison of summarization methods}}
    
    We combine two clustering methods (agglomerative clustering and Leiden community detection) and four summarization methods (BART, T5, DG, and MCI). We set the similarity threshold $\delta = 0.85$ for both clustering methods (Table \ref{tab:sumsum}). We consider news summaries to be ground-truth summaries as the tweets mention these news articles. 
    
    Table~\ref{tab:sumsum} shows that extractive summarization (DG and MCI) ouperforms abstractive summarization (BART and T5). This means most content are repetitions of news content, so the extractive summaries of tweets can be precisely the same as news summaries. However, models for the abstractive summaries try to generate fluent sentences by combining multiple different posts thus are longer and overlap less with news summaries.
    In terms of the abstractive approach, the average summary lengths for Leiden clusters are longer than the agglomerative ones. This is because Leiden generates larger sets of social posts than agglomerative --- around 100 less clusters. This way, Leiden clusters contain more posts, and they are better distributed among the clusters. Therefore, we consider the Leiden method more suitable as it reduces redundancy without losing information, even though the scores of Agglomerative are slightly higher.
    
    \begin{table}
    \caption{Summarization performance comparing different summarization methods}
    \centering
    \resizebox{0.5\textwidth}{!}{
    \begin{tabular}{@{}llllll@{}}
    \hline
    \begin{tabular}[c]{@{}l@{}}Summarization\\ Method\end{tabular} & ROUGE-1         & ROUGE-2         & ROUGE-L         & \begin{tabular}[c]{@{}l@{}}BERT-\\score\end{tabular}       & \begin{tabular}[c]{@{}l@{}}Average \\ Summary\\ Length\end{tabular} \\ \hline
    Agglomerative+BART                                                      & 0.53          & 0.41          & 0.49          & 0.91          & 22.03                                                                              \\
    Agglomerative+T5                                                        & 0.51          & 0.39          & 0.47          & 0.90          & 23.99                                                                                                                             \\
    Agglomerative+DG                                                        & {0.59}          & {0.48}          & {0.56}          & {0.92}          & {21.44}                                                                                                                                \\
    Agglomerative+MCI                                                       & \bf 0.59          & \bf 0.48          & \bf 0.56          & \bf 0.92          & \bf 21.49                                                                                                                                \\ \hline
    Leiden+BART                                                    & 0.50          & 0.38          & 0.47          & 0.91          & 23.44                                                                               \\
    Leiden+T5                                                      & 0.48          & 0.36          & 0.44          & 0.89          & 26.17                                                                                                                               \\
    Leiden+DG                                                      & \textbf{0.59} & \textbf{0.48} & \textbf{0.55} & \textbf{0.92} & \textbf{21.48}                                                                                                                     \\
    Leiden+MCI                                                     & 0.58          & 0.47          & 0.55          & 0.92          & 21.53                                                                                                                              \\ \hline
    \end{tabular}
    }
    \begin{tablenotes}[flushleft]
      \item Note: Average news summary length for Agglomerative and Leiden are 15.20 and 15.99 respectively. Number of clusters for Agglomerative and Leiden are 804 and 705, respectively.
    \end{tablenotes}
    \label{tab:sumsum}
    \end{table}
    
Table~\ref{tab:example} shows an example of summaries for qualitatively illustration of the summarization process for Leiden clustering. All  summaries essentially reduced the redundancy of posts, and capture the central claim of the tweets.
\begin{table*}

\caption{ An example of four summarization results on tweets in one cluster}
\centering
\resizebox{0.99\textwidth}{!}{
\begin{tabular}{@{}p{24cm}@{}}
\hline
\textbf{Original tweets from a cluster (11 out of 241 tweets after removing duplicates and near-duplicates)} \\
\tabitem{Reupping this fact check --\&gt; How false hope spread about hydroxychloroquine to treat covid-19 — and the consequences that followed} \\ 
\tabitem{How Trump’s false hope spread about hydroxychloroquine to treat covid-19 — and the consequences that followed - The Washington Post}\\ 
\tabitem{Coronavirus: How false hope spread about hydroxychloroquine to treat covid-19 - and the consequences that followed |} \\ 
\tabitem{How false hope spread about hydroxychloroquine to treat covid-19 — and the consequences that followed. Four Pinocchios given by the WP.} \\
\tabitem{Well done for " How false hope spread about hydroxychloroquine to treat covid-19 – and the consequences that followed" - highlighted in Journalism Matters survey on Excellence in Reporting Coronavirus}\\ 
\tabitem{How false hope spread about hydroxychloroquine to treat covid-19 — and the consequences that followed Dr. Trump's medicine show: Why is he pushing an unproven drug? Follow the money}\\ 
\tabitem{For all you MAG Ats who keep pushing the lie: " How false hope spread about hydroxychloroquine to treat covid-19 — and the consequences that followed" maga KAG Unscientific Pontificator M Dfrom Trump University}\\ 
\tabitem{Trump is making baseless, irresponsible medical recommendations based on rumor and social media idiocy. Analysis | How false hope spread about hydroxychloroquine to treat covid-19 — and the consequences that followed}\\ 
\tabitem{How false hope spread about hydroxychloroquine to treat covid-19 — and the consequences that followed" excellent \&amp; important Fact Checker, which also explains how social media gave this dangerous info undeserved oxygen} \\ 
\tabitem{Trump and his enablers pushing dangerous and unproven medical advice as if their doctors. Should be a law against this. How false hope spread about hydroxychloroquine to treat covid-19 — and the consequences that followed}\\ 
\tabitem{The only thing fake is you. Fake President How false hope spread about hydroxychloroquine to treat covid-19 — and the consequences that followed}\\ \hline
\textbf{BART summarization}                                   \\
How false hope spread about hydroxychloroquine to treat covid-19 – and the consequences that followed. Four Pinocchios given by the WP.                               \\ \hline
\textbf{T5 summarization}                     \\
"how false hope spread about hydroxychloroquine to treat covid-19 – and the consequences that followed" "why is he pushing an unproven drug? follow the money" "trump is making baseless, irresponsible medical recommendations based on rumor"                                     \\ \hline
\textbf{DG summarization}                                 \\
How false hope spread about hydroxychloroquine to treat covid-19 — and the consequences that followed - msnNOW                                                                  \\ \hline
\textbf{MCI summarization}                                 \\
Is mystery How false hope spread about hydroxychloroquine to treat covid-19 — and the consequences that followed \\ \hline
\textbf{News Summary (Gold)}                               \\
President Trump has repeatedly touted the anti-malarial medications hydroxychloroquine and chloroquine as the much-needed solution to COVID-19                   \\ \hline
\end{tabular}%
}
\label{tab:example}
\end{table*}

\subsubsection{\textbf{Analysis of the graph of summaries}}

\begin{figure}[t]
        \centering
            \includegraphics[width=0.37\textwidth]{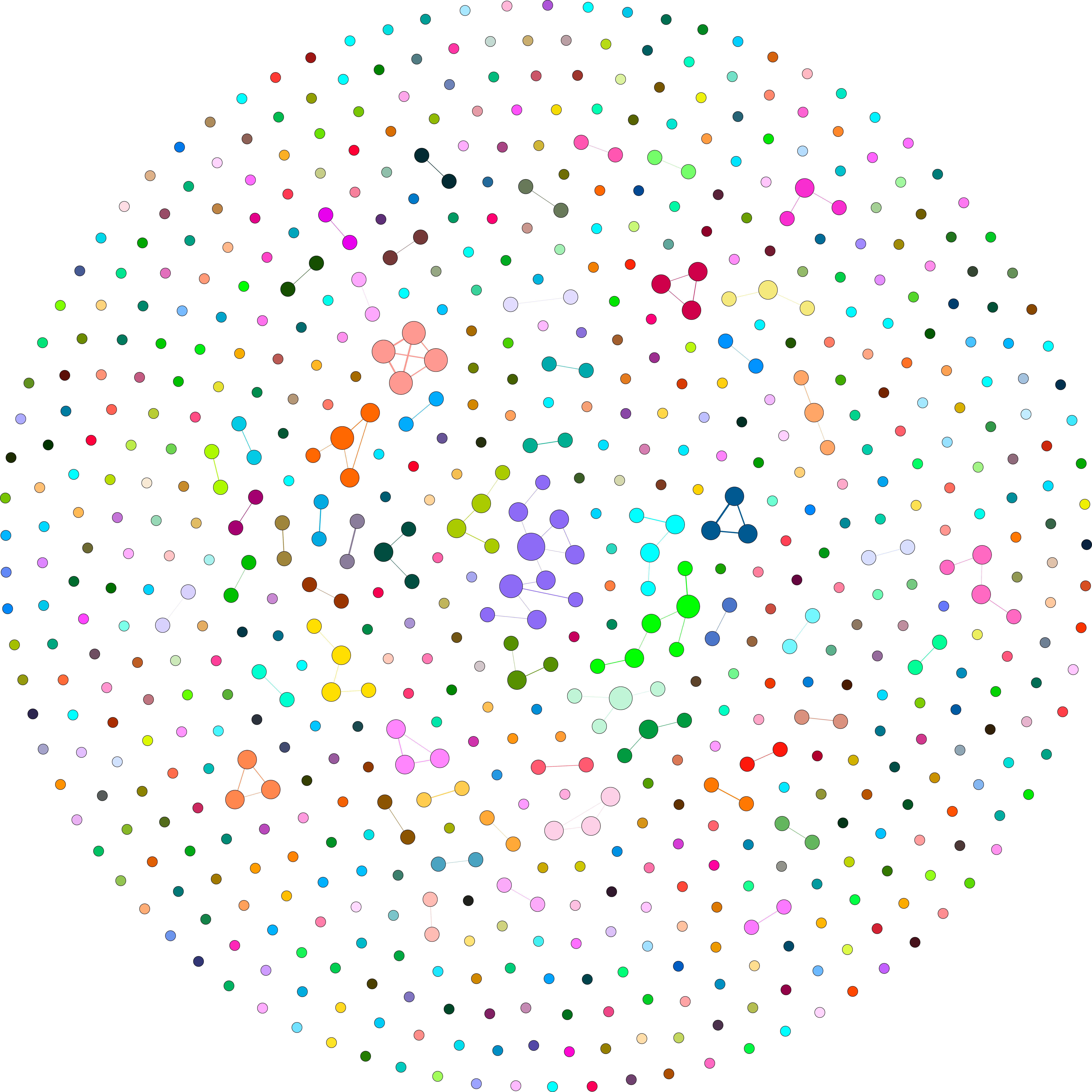}
            \caption{The communities from the graph of summaries generated by BART. Each connected component is a community.
            }
    \label{f:grasum}
    \end{figure}

To validate the robustness of our clustering and summarization methods, we construct a graph of summaries to see if there are similar summaries. We take the summaries generated from \textbf{Leiden+BART} and perform a Leiden community detection with similarity threshold of 0.75, then visualize the communities in the graph. We show the graph in Figure \ref{f:grasum}. In this graph, there are 609 communities. We can see that it is a sparse graph; only a few communities have more than one node, indicating the clustering and summarization effectiveness. 

To further check if the summaries in the same community are similar, we show two examples of all the summaries in a community for communities 1 and 2 (communities are sorted in descending order with the number of summaries) in Table \ref{tab:sumcom}. We can see that the summaries in one community are similar to each other and related to similar topics, but they are not related to one specific claim. Therefore we conclude that our clustering and summarization find reasonable and useful claims and a further reduction would risk losing information.

\begin{table*}
\caption{Examples of all BART summaries in a community of graph of summaries (community 1 and community 2)}
\centering
\resizebox{0.99\textwidth}{!}{
\begin{tabular}{@{}p{26cm}@{}}
\hline
\textbf{All summaries in Community 1}                       \\
\tabitem{What’s a coronavirus superspreader?}\\ 
\tabitem{How does the coronavirus work?}\\ 
\tabitem{People with coronavirus may be most infectious in the first week of symptoms. SARS CoV2 COVID19.}\\ 
\tabitem{COVID-19: Information on symptoms, transmission – Mayo Clinic News Network. Covid19 Corona Virus CoronavirusUSA: Terms to know.}\\ 
\tabitem{GI Symptoms and Coronavirus (COVID-19) from. GI Symptoms of CO VID-19 from.}\\ 
\tabitem{Coronavirus: How does the Covid-19 alert level system work?}\\ 
\tabitem{What are the early symptoms of coronavirus (COVID-19)?}\\ 
\tabitem{People with coronavirus may be most infectious in the first week of symptoms. That could lend more weight to the argument in favor of wearing a mask while in public.}\\ 
\tabitem{Here are answers to key questions about the virus, including how to protect yourself and what to expect. What questions do you have about the new coronavirus?} \\
\tabitem{Get an answer about the coronavirus, how does it kill, truth about masks, do they work, are pets safe, do HVAC systems spread the coronvirus, do quarantine’s work, what about cures, vaccines, treatment, how long will this coronav virus last and more.} \\ \hline
\textbf{All summaries in Community 2}             \\
\tabitem{COVID-19 can be spread by people who do not have symptoms and do not know that they are infected. CDC recommends that you wear masks in public settings around people who don't live in your household and when you can’t stay 6 feet away from others.} \\
\tabitem{New Evidence Shows Wearing Face Mask Can Help Coronavirus Enter the Brain and Pose More Health Risk, Warn Expert. He stresses that only ill people should wear face masks.}\\ 
\tabitem{The CDC recommends wearing a cloth face mask in public to help slow the spread of coronavirus. But the evidence for the efficacy of surgical or homemade masks is limited, and masks aren't the most important protection.}\\ 
\tabitem{Dr. Russell Blaylock warns that not only do face masks fail to protect the healthy from getting sick, they also create serious health risks to the wearer.}\\
\tabitem{The CDC does not recommend that asymptomatic,healthy people wear a facemask to protect themselves from respiratory diseases. Facemasks should be used by people who show symptoms of COVID-19 to help prevent the spread of the disease to others.}\\
\tabitem{The CDC does not recommend that people who are healthy wear facemasks. It does recommend that those who are not healthy wear them.}\\ \hline
\end{tabular}
}
\label{tab:sumcom}
\end{table*}

\subsection{Human-in-the-Loop}
We invite fact-checker journalists to evaluate the summarization methods, evaluating the four proposed methods. We also include the news summary (ground-truth summary) in the comparison. We asked the specialist to give a score ranging from 1-5 (one means the summary is not representative for the posts in the cluster, and five means the summary is very representative for the posts). We randomly select 50 clusters for each summarization method; the clustering method is Leiden. We average the scores along 50 clusters, and the average scores for BART, T5, DG, MCI and news summaries are: 4.90, 4.92, 4.96, 4.96 and 4.68 respectively.

We can see that overall all summarization methods have an average score higher than 4, which means they are highly representative. Extractive methods' scores are slightly higher than abstractive ones as the latter sometimes bring additional comments, which are often wrong or are prejudiced. Surprisingly, the news summary, which we treat as ground-truth, has the lowest average score according to human evaluation. Specifically, some summaries received low scores because they do not offer sufficient information to obtain the claim in question or related to a similar but different claim.

\section{Conclusion}

While automated fact-checking solutions are not near ready for deployment in real-world scenarios, it is key important to assist human checkers to improve speed and comprehensive inspection. This paper fills the gap between manual and automated fact-checking through a two-step pipeline: grouping similar messages together and summarizing them into one claim, which a human will then check. We test our pipeline by combining two clustering and four summarization methods. The results show that the framework can largely reduce the number of original social media posts in more than 97\%  --- from 28,818 tweets to 700 summary claims --- and deliver more informative claims that enrich the knowledge about the clustered messages for the fact-checking process.

This work is an initial step toward more efficient and effective fact-checking with human-in-the-loop. However, we still face some challenges: 1) we do not have reliable ground-truth labels for the clustering stage and oracle summaries for the summarization evaluation; 2) the clustering methods require calculating an entire similarity matrix, which can be a bottleneck for scalable fact-checking; 3) some summaries are not claims thus may not be of interest to fact-checkers. In future work, we plan to build a more standard dataset for evaluation. We also plan to apply this pipeline on more diverse datasets with hundreds of thousands or even millions of posts to address the importance of clustering and summarizing efficiency. Another possible direction is to refine the summaries by removing non-claims to further reduce their number.

\section*{Acknowledgments}
Research funded by Fapesp -- Grant D\'ej\`aVu \#2017/12646-3.

\bibliographystyle{IEEEtran}
\bibliography{IEEEabrv,ref}

\end{document}